\pdfoutput=1

\documentclass[11pt]{article}

\usepackage[final]{acl}

\usepackage{times}
\usepackage{latexsym}

\usepackage[T1]{fontenc}

\usepackage[utf8]{inputenc}

\usepackage{microtype}

\usepackage{inconsolata}

\usepackage{hyperref}
\usepackage{url}
\usepackage{graphicx} 
\usepackage{booktabs}
\usepackage{wrapfig}
\usepackage{tabularx}
\usepackage{subfigure}
\usepackage{textcomp}
\usepackage{array}
\usepackage{caption}
\usepackage{multirow}
\usepackage{titlesec}
\usepackage{setspace}
\usepackage{algorithm}
\usepackage{algorithmic}
\usepackage{appendix}
\usepackage{listings}
\usepackage{color, colortbl}
\usepackage{amsmath}
\usepackage{amsfonts}
\usepackage{enumitem}
\usepackage{etoolbox}
\usepackage{stfloats}
\usepackage{booktabs} 
\usepackage{subcaption}
\usepackage{tabularray}
\usepackage{makecell}
\usepackage{lipsum}
\usepackage{tablefootnote}
\usepackage{tcolorbox}

\newtcolorbox{titleEnv}{
colframe=black!80,
colback=gray!10,
fonttitle=\bfseries,
coltitle=black,
left=3pt,
right=3pt,
top=3pt,
bottom=3pt,
boxrule=0.4mm,
arc=3mm
}

\title{LoRAMoE: Alleviate World Knowledge Forgetting in Large \\ Language Models via MoE-Style Plugin}

\author{\textbf{Shihan Dou}$^{1}\thanks{{ }{ }Equal contribution.}$, 
\textbf{Enyu Zhou}$^{1 *}$, 
\textbf{Yan Liu}$^{1}$, 
\textbf{Songyang Gao}$^{1}$, 
\textbf{Jun Zhao}$^{1}$, 
\textbf{Wei Shen}$^{1}$, 
\\
\textbf{Yuhao Zhou}$^{1}$, 
\textbf{Zhiheng Xi}$^{1}$\textbf{,} 
\textbf{Xiao Wang}$^{1}$\textbf{,}
\textbf{Xiaoran Fan}$^{1}$\textbf{,}
\textbf{Shiliang Pu}$^{2}$\textbf{,} 
\textbf{Jiang Zhu}$^{2}$\textbf{,}
\\
\textbf{Rui Zheng}$^{1}$\textbf{,} 
\textbf{Tao Gui}$^{1}$\thanks{{ }{ }Corresponding author.}\textbf{\ ,} 
\textbf{Qi Zhang}$^{1\dag}$\textbf{,} 
\textbf{Xuanjing Huang}$^{1}$
\\
$^{1}$ NLP Group, Fudan University\\
$^{2}$ Hikvision Inc \\
\texttt{shdou21@m.fudan.edu.cn, eyzhou23@m.fudan.edu.cn}\\
\texttt{\{rzheng20, tgui, qz\}@fudan.edu.cn}
}

\begin{document}
\maketitle
\begin{abstract}
Supervised fine-tuning (SFT) is a crucial step for large language models (LLMs), enabling them to align with human instructions and enhance their capabilities in downstream tasks.
Increasing substantially instruction data is a direct solution to align the model with a broader range of downstream tasks or notably improve its performance on a specific task.
However, we find that large-scale increases in instruction data can damage the world knowledge previously stored in LLMs.
To address this challenge, we propose LoRAMoE, a novelty framework that introduces several low-rank adapters (LoRA) and integrates them by using a router network, like a plugin version of Mixture of Experts (MoE).
It freezes the backbone model and forces a portion of LoRAs to focus on leveraging world knowledge to solve downstream tasks, to alleviate world knowledge forgetting.
Experimental results show that, as the instruction data increases, LoRAMoE can significantly improve the ability to process downstream tasks, while maintaining the world knowledge stored in the LLM \footnote{\ \url{https://github.com/Ablustrund/LoRAMoE}}.

\end{abstract}

\begin{figure}
    \centering
    \includegraphics[width=0.48\textwidth]{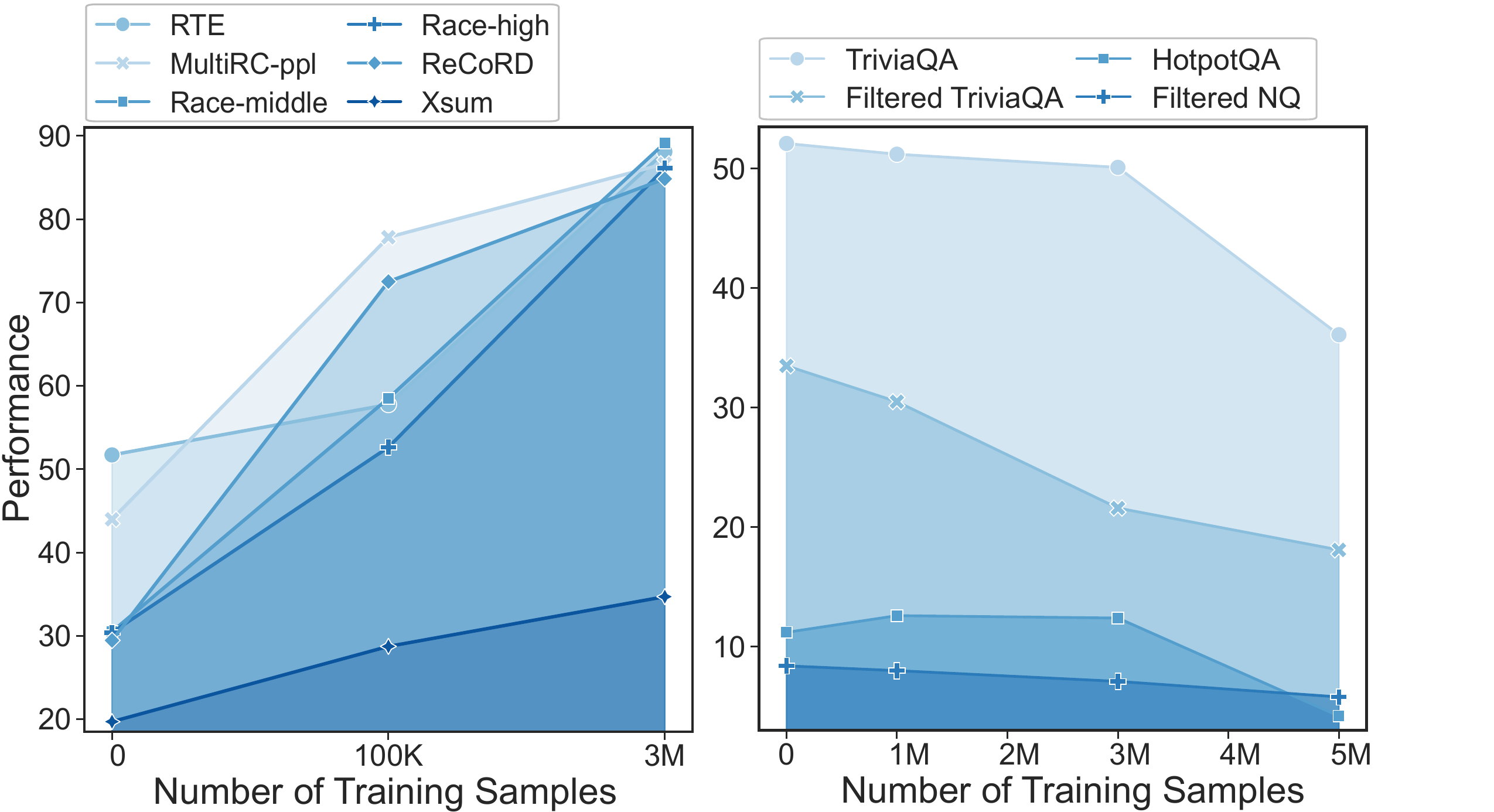}
    \caption{\textbf{(Left)} With the number of fine-tuning data increases from 10K to 3M, the performance of many downstream tasks is significantly improved.
    \textbf{(Right)} With the amount of instruction data increasing, fine-tuning the language models results in a decline in performance on the benchmarks that measure their world knowledge, such as TriviaQA \citep{han2019episodic}, Natural Questions \citep{kwiatkowski2019natural}. 
    The details of training implementation can be seen in Section~\ref{sec:sft}.
    }
    \label{fig:firstpage}
    \vspace{-1.5em}
\end{figure}

\section{Introduction}

Supervised fine-tuning (SFT) provides a pivotal technique to make large language models (LLMs) follow human instructions and improve their performance of downstream tasks \citep{chung2022scaling,ouyang2022training}. 
Although some studies \citep{zhou2023lima, cao2023instruction} indicate that LLMs trained on a little data can follow instructions well, increasing the amount of data is a straightforward way to enhance their ability to multiple downstream tasks or improve their performance on a specific task, as shown in the left of Figure~\ref{fig:firstpage}.



However, the large-scale increase in instruction data can destroy the world knowledge stored in LLMs, as illustrated in the right of Figure~\ref{fig:firstpage}. 
Specifically, as the amount of instruction data increases, we observe a notable decline in performance on Closed-Book Question Answering (CBQA) datasets, which are used to measure world knowledge in LLMs \cite{touvron2023llama, neeman2022disentqa}.
In the paradigm of supervised fine-tuning, the conflict between maintaining world knowledge inside LLMs and improving their performance on downstream tasks by scaling up instruction data has not been thoroughly examined.

\begin{figure*}
    \centering
    \includegraphics[width=\textwidth]{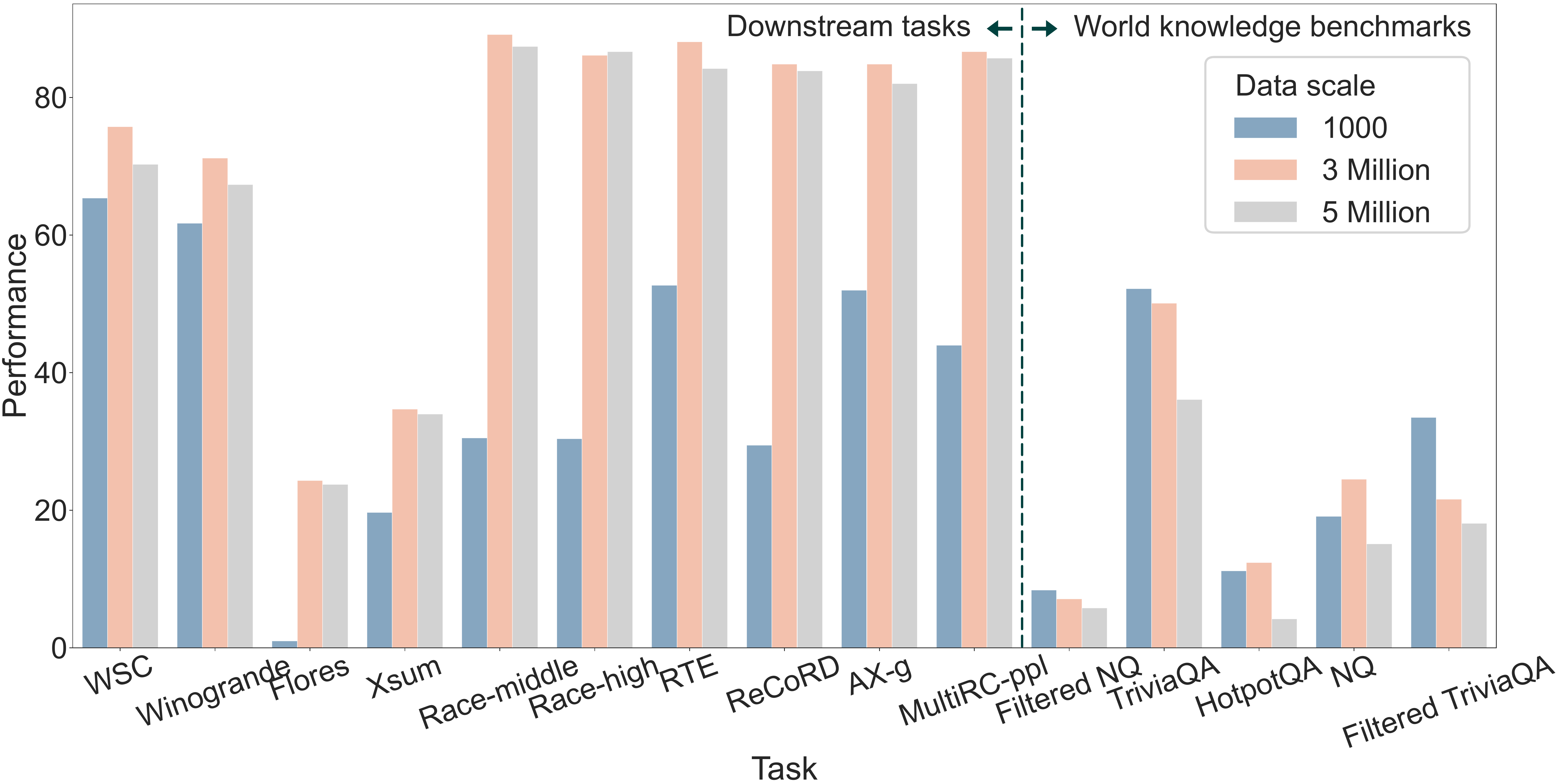}
    \caption{Performance on the various tasks after expanding the amount of fine-tuning data. For most of the downstream tasks (e.g., NLI and summarization), with the expansion of training data, performance on these tasks remains stable after improvement.
    Whereas, for the world knowledge benchmark, a significant \textbf{decline} can be witnessed after a large amount of instruction data.}
    \label{fig:conflict-mainresult}
    \vspace{-0.5em}
\end{figure*}

In this paper, we propose LoRAMoE, a novelty framework for SFT, to enhance the models' capability of solving downstream tasks, while alleviating world knowledge forgetting during the training phase.
LoRAMoE is a Mixture-of-Experts-style (MoE-style) plugin, which introduces several low-rank adapters (LoRA \cite{hu2021lora}) as experts and integrates them by using a router network.
The router network automatically assigns weights to experts, which can improve the LLM's performance on multiple downstream tasks.

To demonstrate the efficacy of our proposed method, we conduct extensive experiments across a range of downstream tasks.
Experiment results show that LoRAMoE can significantly improve LLM's ability to address the various downstream tasks by fine-tuning the model on a large amount of instruction data, while maintaining the world knowledge stored in the model.
In addition, we further evaluate our method by visualizing the expert weight for tasks. 
The result indicates that LoRAMoE adequately alleviates world knowledge forgetting and achieves an improvement of models by fostering collaboration among experts.
The main contributions of our paper are as follows:
\begin{enumerate}
    \item We find that significantly increasing the amount of instruct data during the SFT phase can damage the world knowledge inside the LLMs. 
    The need for improvement in downstream tasks by scaling up instruction data conflicts with maintaining the world knowledge inside the model.
    \item We introduce LoRAMoE, a novelty framework for SFT, which introduces LoRAs as experts and integrates them by the router. LoRAMoE can enhance the model's ability to address downstream tasks, while alleviating the world knowledge forgetting.
    \item Extensive Experiments demonstrate the efficacy of our proposed approach in multi-tasks and mitigating the forgetting of world knowledge inside the model. 
    The visualizing experiment shows that LoRAMoE can achieve an improvement by fostering collaboration among experts.
\end{enumerate}

\section{Motivation}
\label{sec:conflict}


In this section, we verify that a large-scale SFT can cause irreversible damage to world knowledge within the LLMs while improving the LLMs' performance in various downstream tasks.
\subsection{A Diverging Trend}
\label{sec:sft}
We constructed a dataset containing seven categories of tasks with a total of five million training samples, and used it to conduct SFT on a Llama-2-7B model.
The implementation details are described in Appendix~\ref{appendix:SFT-settings}.
During the expansion of fine-tuning data, we observed a diverging trend in the performance across two types of tasks, as shown in Figure~\ref{fig:conflict-mainresult}:

Across downstream tasks such as summarization, Natural Language Inference (NLI), machine translation, and others, the performance of the fine-tuned model initially showed a magnificent increase and eventually stabilized at a promising level. 
However, when it comes to closed-book QA (CBQA) tasks that are used as world knowledge benchmark \citep{touvron2023llama,neeman2022disentqa}, the model's performance catastrophically declines under the baseline Notably, with the training data expanding, a contiguous decline can be witnessed. Moreover, this decline will occur earlier if the test set is filtrated. Appendix~\ref{sec-1kw} case with a larger dataset including more tasks shows an even steeper drop on world knowledge benchmarks, although performance remains competitive on others.




\subsection{The Irreversible Knowledge Forgetting}

In this section, we dissect the reason behind the decline on these world knowledge benchmarks during the expansion of fine-tuning data. We find this results from the occurrence of irreversible knowledge forgetting inside the LLM.
\begin{figure}
  \centering
  \includegraphics[width=0.46\textwidth]{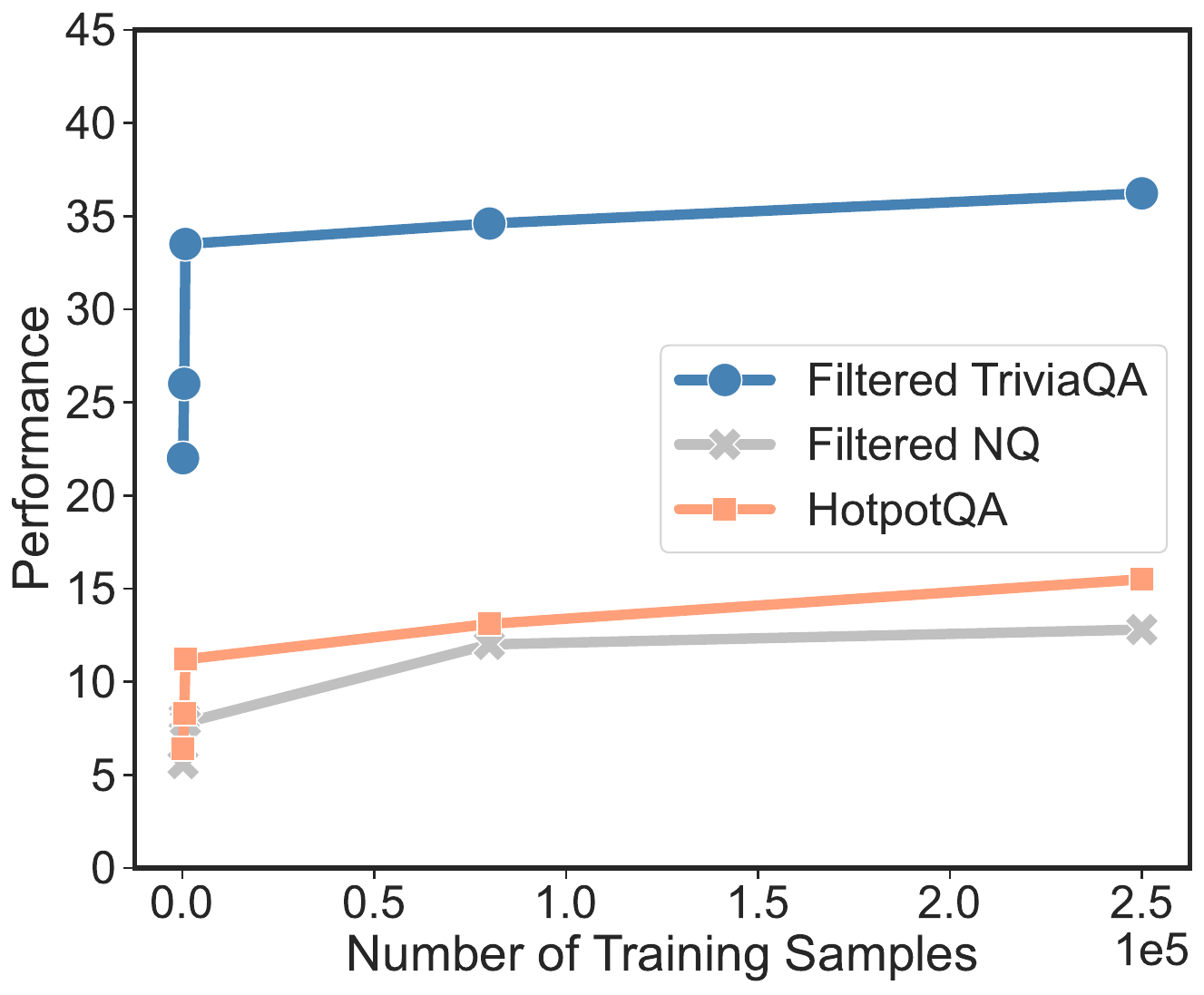} 
      \caption{Performance on world knowledge benchmarks after training on CBQA solely. Its performance rises greatly after training with very few samples and remains relatively stable thereafter.}
    \label{fig:conflict-kn}
    \vspace{-0.5em}
\end{figure}

\textbf{
The performance on world knowledge benchmarks highly relies on the knowledge and skills learned during pre-training phase.
} 
To investigate the relationship between the performance on world knowledge benchmarks and the knowledge embedded in pre-trained models \citep{petroni2019language, roberts2020much,alkhamissi2022review}, we conduct fine-tuning solely on the CBQA dataset with 250k samples and run evaluation on the test sets without train-test overlap. Results in Figure \ref{fig:conflict-kn} show initial training boosts performance significantly, especially the first 1\% (approximately 1k samples), with limited gains thereafter. This is because early fine-tuning aligns existing knowledge with new instructions, improving CBQA results. However, due to minimal training-testing data overlap, adding more samples doesn't further enhance performance. Thus, a model's benchmark success relies on world knowledge acquired from the pre-training. 

Given this, it is naturally assumed that \textbf{the diminished performance on knowledge benchmark stems from the damage of knowledge stored in the LLM due to large-scale instruction tuning}.
To verify the hypothesis, we sequentially fine-tuned a model using two datasets, first excluding CBQA data, then with CBQA data. Results presented in Table~\ref{tab:conflict-continue} show a great decline in knowledge capabilities versus the original LLM. 
This indicates that the world knowledge within the model was compromised during the first stage of large-scale fine-tuning, resulting in the model's inability to forge the alignment between human instructions and the already destroyed knowledge in the subsequent stage of fine-tuning solely with CBQA.

To sum up, the pursuit of enhancing performance on downstream tasks through the expansion of training data conflicts the preservation of world knowledge within the model in vanilla SFT.

\begin{table}
  \centering
  \resizebox{0.48\textwidth}{!}{
  \begin{tabular}{c|c|c|c}
    \toprule
    \toprule
    \textbf{Task Name} & \textbf{Baseline} & \textbf{\begin{tabular}[c]{@{}c@{}}SFT solely\\ on CBQA\end{tabular}} & \textbf{\begin{tabular}[c]{@{}c@{}}Two-stage \\ Fine-tuning\end{tabular}} \\ 
    \midrule
    \textbf{TriviaQA} & 33.5 & 36.22 & 13.7 \\
    \textbf{NQ} & 7.8 & 12.8 & 3.6 \\
    \textbf{HotpotQA} & 11.2 & 16.1 & 7.1 \\ 
    \bottomrule
    \bottomrule
  \end{tabular}
  }
  \caption{
  Performance from left to right: LlaMA-2-7B, model tuned on CBQA, and model tuned on 3M instructions then on CBQA. Despite further tuning on CBQA, the large-scale SFT model's knowledge-answering doesn't improve, staying below the baseline.
  }
  \label{tab:conflict-continue}
  \vspace{-0.5em}
\end{table}



\begin{figure*}
    \centering
    \includegraphics[width=\textwidth]{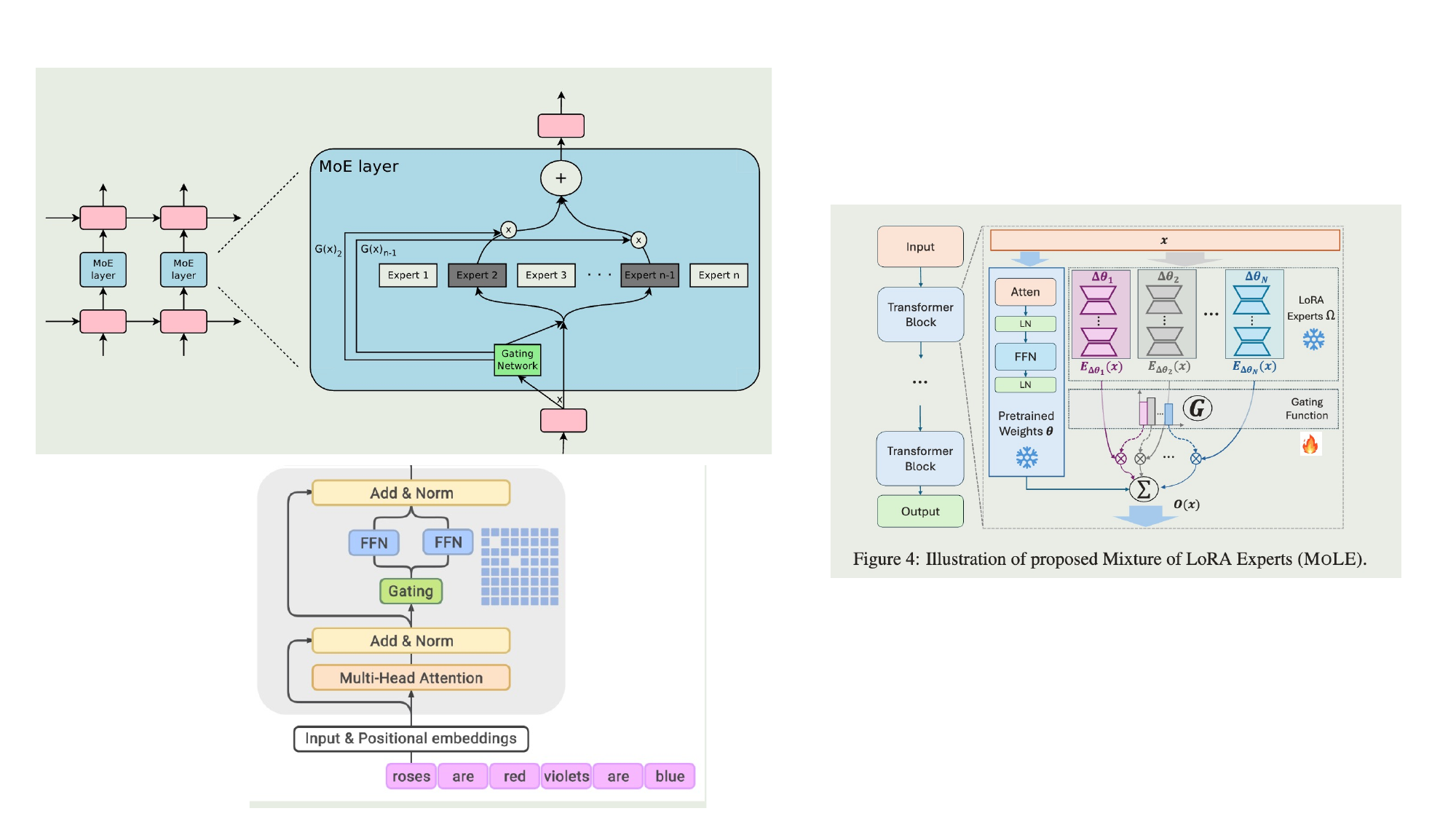}
    \caption{The architecture of LoRAMoE, compared with classic MoE. LoRAMoE utilizes multiple LoRAs as adaptable experts and a router to gate them in the FFN layer of every transformer block. During the training process, only the experts and the router are optimized. }
    \label{fig:method-architecture}
    \vspace{-0.5em}
\end{figure*}

\section{LoRAMoE}
In this section, we elaborate on the methodological details of LoRAMoE, which is an MoE-style plugin and introduced Localized Balancing Constraint during the training phase to alleviate the world knowledge, as shown in Figure~\ref{fig:method-architecture}.

\subsection{Architecture}

The left of Figure~\ref{fig:method-architecture} illustrates the forward process of the standard MoE architecture \citep{shazeer2016outrageously,switchtransformers,gshard}.
In the MoE, the router assigns weights of experts according to the data, allowing them to divide their labor to complete the forward process \citep{moe1991first}.
The key sight of LoRAMoE is that we freeze the backbone model to maintain world knowledge and introduce experts to leverage this knowledge to address tasks, while improving the performance on multiple downstream tasks.  
Additionally, we utilize the LoRA \cite{hu2021lora} as the architecture of the expert to improve training and inference efficiency.


Formally, for the traditional transformers architecture, the forward propagation process of the feed-forward neural (FFN) network block can be simplified as follows:
\begin{equation}
\small
f(x) = x + f_{\text{FNN}}(x).
\end{equation}
The matrix operation of the linear layer in this forward propagation can be expressed as:
\begin{equation}
\small
o = Wx = W_0x + \Delta Wx
\end{equation}
where $W_0 \in \mathbb{R}^{d_{\text{in}} \times d_{\text{out}}}$ represents the parameter matrix of the backbone model and $\Delta W \in \mathbb{R}^{d_{\text{in}} \times d_{\text{out}}}$ denotes the updated parameter during the training phase. 
For LoRAMoE, we replace the linear layer in the FFN block with the MoE-style plugin, which makes experts collaborate to address tasks.
During the training phase, we freeze the backbone to maintain the world knowledge and only update $\Delta W$. 
Consider the LoRAMoE layer containing $N$ experts, which is denoted as $\{E_i\}_{i=1}^{N}$, the forward process of the layer can be mathematically expressed as follows: 
\begin{equation}
\small
o = W_0x + \Delta Wx = W_0x + \sum_{i=1}^{N}G(x)_i E_i(x)
\end{equation}
where $E_i(\cdot)$ and $G(\cdot) = Softmax(xW_{g})$ represent the $i$-th expert and the router in the LoRAMoE layer, respectively.
The $W_{g}$ is the trainable parameter matrix of the route network.
By this, the experts and the outer work in tandem, enabling the experts to develop varied capabilities and efficiently handle diverse types of tasks.

In addition, LoRA has been proven to be both effective and efficient for the SFT phase of LLMs \citep{wang2023orthogonal, liu2022few, pan2022st}. 
To enhance the efficiency and resource conservation of the fine-tuning process, we replace the parameter matrix of the experts with a low-rank format.
Specifically, the matrix $\Delta W_E \in \mathbb{R}^{d_{\text{in}} \times d_{\text{out}}}$ of the expert $E(\cdot)$ in the LoRAMoE layer can be written as follows:
\begin{equation}
\small
    \Delta W_E = BA
\end{equation}
where $A \in \mathbb{R}^{d_{\text{in}} \times r}$, $B \in \mathbb{R}^{r \times d_{\text{out}}}$, and the rank $r \ll \min(d_{\text{in}}, d_{\text{out}})$.
LoRA contributes to a significant reduction in the trainable parameters, thereby enhancing efficiency and saving costs during the fine-tuning process.

Overall, the forward process of the LoRAMoE layer replaced the traditional FFN layer can be represented as:
\begin{equation}
\small
        o = W_0x + \frac{\alpha}{r} \sum_{i=1}^{N}\omega_i \cdot B_i A_i x
\end{equation}
where $\omega_i$ denotes the weight of $i$-th expert and $\alpha$ is the constant hyper-parameter, approximately equivalent to the learning rate.



\begin{figure}
    \centering
    \includegraphics[width=0.46\textwidth]{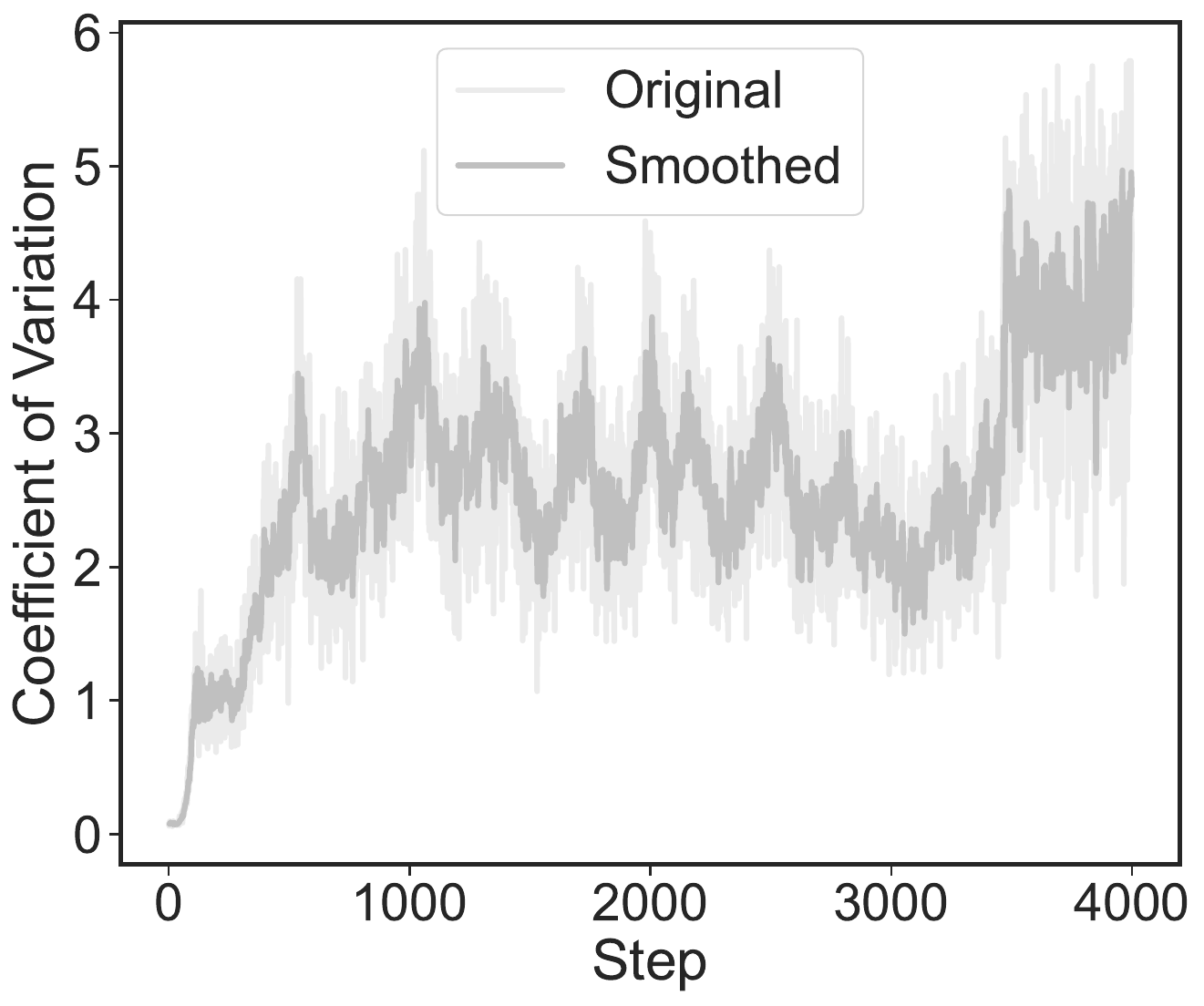}
    \caption{The coefficient of variation for the experts of the unconstrained LoRAMoE progressively escalates and sustains at a high value, i.e., approximately three, similar to the phenomenon observed at \citet{shazeer2016outrageously}.
    This indicates that the router assigns large weights to the same few experts.
    }
    \label{fig:imbalance_loss}
\end{figure}

\subsection{Localized Balancing Constraint}

The imbalance of the experts' utilization is a typical problem in MoE \citep{shazeer2016outrageously, switchtransformers}, which is also observed in our proposed method, as shown in Figure~\ref{fig:imbalance_loss}. 
The conventional solution is balancing expert utilization \citep{shazeer2016outrageously}, which involves making the coefficient of variation of the experts' importance as the loss function. 
However, this method assumes all the training samples are under the same distribution, which ignores the fact that samples may be from different distributions such as the question-answering task and other downstream tasks, more detailed analysis and conceptual proof in Appendix~\ref{app-balance}.

\begin{figure}
    \centering
    \includegraphics[width=0.46\textwidth]{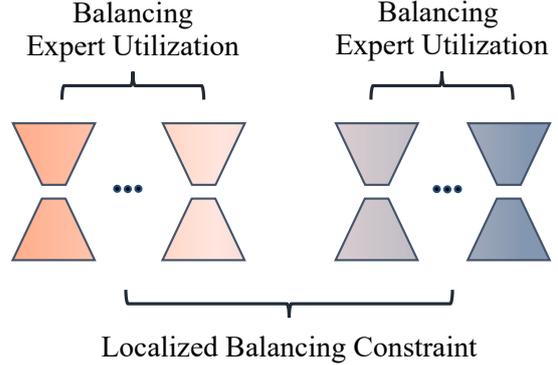}
    \caption{
    Localized balancing constraint. 
    We softly force experts to focus on two types, one for leveraging world knowledge by learning on its related tasks, and another for concentrating on other downstream tasks.
    Meanwhile, the experts in solving the same aspect are balancing.
    }
    \label{fig:our_method}
\end{figure}

Considering the mixed characteristics of data distributions are important, during the training phase, we introduce localized balancing constraint, a novelty balancing expert utilization method to make a portion of experts focus more on leveraging world knowledge to solve tasks.
As shown in Figure~\ref{fig:our_method}, during the fine-tuning phase, we softly constrain experts to concentrate on two aspects, one of which focuses on leveraging world knowledge by learning on its related datasets, while another focuses on other downstream tasks.
In addition, all experts within the same aspects are balanced such as balancing expert utilization.


Formally, we define the importance matrix $\mathbf{Q}$ of the LoRAMoE layer and $\mathbf{Q}_{n, m}$ denotes the sum of router values of the $n$-th expert for the $m$-th training sample in a batch, which can be represented as follows:
\begin{equation}
\small
    \mathbf{Q}_{n, m} = \sum_{j=1}^{T_m}G(x_j)_i = \frac{\exp(\omega_i^j / \tau)}{\sum_{k=1}^{N}\exp(\omega_i^j / \tau)}
\end{equation}
where $N$ and $T_m$ denote the number of experts and the number of tokens of $m$-th training sample, respectively. 
$x_j$ is the hidden input of the $j$-th token.
We then define the coefficient matrix $\mathbf{I}$ with the same size of $\mathbf{Q}$, corresponding to the importance matrix $\mathbf{Q}$.  
$\mathbf{I}_{n, m}$ denotes the importance coefficient of $\mathbf{Q}_{n, m}$, which can be written as follows:
\begin{equation}  
\small
\mathbf{I}_{n, m} = 
\left\{  
    \begin{array}{lr}  
             1+\delta, & \text{Type}_e(n) = \text{Type}_s(m) \\  
             1-\delta, & \text{Type}_e(n) \neq \text{Type}_s(m)
    \end{array}  
\right.  
\end{equation}  
where $\delta \in [0, 1]$ controls the degree of imbalance between experts types. 
$\text{Type}_e(n)$ and $\text{Type}_s(m)$ are pre-defined target type of $n$-th expert and the task type of $m$-th training sample in a batch, respectively.

We categorize the instruction data into two distinct types: world knowledge-related tasks such as TriviaQA, and other downstream tasks such as Flores.
Then, we enable a portion of experts to learn on world knowledge-related tasks to align human instructions with world knowledge, while making other experts focus more on enhancing the performance of downstream tasks.
Formally, suppose that $\mathbf{I}_{i,k}$ and $\mathbf{I}_{j,k}$ denote the importance coefficient of the $i$-th and $j$-th expert for the $k$-th sample, respectively.
If experts are in the same group, their values at corresponding positions in the coefficient matrix are identical, i.e., $\mathbf{I}_{i,k} = \mathbf{I}_{j,k}$.
This indicates that these experts have the same importance because they are assigned to focus on learning the same type of tasks.
On the contrary, the values of experts from distinct groups at their coefficient matrix are different, i.e., $\mathbf{I}_{i,k} \neq \mathbf{I}_{j,k}$.


The localized balancing constraint loss $\mathcal{L}_{lbc}$ is defined to measure the dispersion of the weighted importance matrix $\mathbf{Z} = \mathbf{I} \circ \mathbf{Q}$, which can be mathematically represented as:
\begin{equation}
\small
    \mathcal{L}_{lbc} = \frac{\sigma^2(\mathbf{Z})}{\mu(\mathbf{Z})}
\end{equation}
where $\sigma^2(\mathbf{Z})$ and $\mu(\mathbf{Z})$ represent the variance and mean of $\mathbf{Z}$, respectively.
Specifically, if a specific sample is from the world knowledge-related dataset, experts focusing on solving this type will have larger values in the coefficient matrix $\mathbf{I}$.
Optimizing the loss $\mathcal{L}_{lbc}$ reducing can make corresponding experts learn more from this sample and be assigned a larger weight by the router.
Meanwhile, experts solving the same type of task are balanced such as \citet{shazeer2016outrageously}.
In addition, the constraint is soft to encourage cooperation among experts to preserve the capacity for generalization.




Overall, localized balancing constraint $\mathcal{L}_{lbc}$ achieves a localized balance between two types of experts: one specializes in leveraging world knowledge by training more on world knowledge-related datasets, while the other concentrates on various downstream tasks.
The loss of LoRAMoE can be represented as follows:
\begin{equation}
\small
    \mathcal{L}_{\text{total}} = \mathcal{L} + \beta \mathcal{L}_{lbc}
\end{equation}
where $\mathcal{L}$ is the next-token prediction loss of LLMs and $\beta$ controls the strength of localized balancing constraint.
In the training phase, we freeze the backbone model and the trainable parameters are only those of the experts and routers within the LoRAMoE layers. 
In the inference process, the router automatically assigns weights to all experts, which avoids the need for pre-specified data types.

\begin{table*}[htbp]
  \centering
  \begin{spacing}{0.85}
    \setlength{\tabcolsep}{2.5mm}{
\begin{tabular}{c|cc|cc|cc}
\toprule
\toprule
\textbf{Task Name} & \textbf{Baseline} & \textbf{\begin{tabular}[c]{@{}c@{}}SFT solely on\\       CBQA\end{tabular}} & \textbf{SFT} & \textbf{LoRA} & \textbf{LoRAMoE} & \textbf{\begin{tabular}[c]{@{}c@{}}LoRAMoE \\      (with $\mathcal{L}_{lbc}$)\end{tabular}} \\
\midrule
\textbf{WSC} & 65.4 & - & \textbf{76.0} & 65.4 & 71.2 & 70.2 \\
\textbf{winogrande} & 61.7 & - & \textbf{71.2} & 64.3 & 66.3 & 69.6 \\
\textbf{Flores} & 0.1 & - & 24.3 & \textbf{26.6} & 26.4 & 25.9 \\
\textbf{Xsum} & 19.7 & - & 34.7 & 34.5 & \textbf{34.8} & 33.2 \\
\textbf{Race-middle} & 30.5 & - & 89.1 & 78.8 & 84.5 & \textbf{90.0} \\
\textbf{Race-high} & 30.4 & - & 86.1 & 75.3 & 80.6 & \textbf{86.5} \\
\textbf{RTE} & 52.7 & - & \textbf{88.1} & 77.3 & 80.9 & 87.4 \\
\textbf{ReCoRD} & 29.4 & - & 84.8 & 83.2 & 84.3 & \textbf{85.9} \\
\textbf{AX-g} & 52.0 & - & 84.8 & 76.1 & 81.7 & \textbf{87.1} \\
\textbf{multiRC} & 44.0 & - & 86.7 & 81.4 & 87.3 & \textbf{87.9} \\
\midrule
\textbf{TriviaQA} & 52.2 & 57.8 & 51.1 & 47.8 & 55.3 & \textbf{58.1} \\
\textbf{NQ} & 18.5 & 28.6 & 24.5 & 16.2 & 23.8 & \textbf{28.0} \\
\textbf{Filtered TriviaQA} & 33.5 & 36.2 & 21.6 & 33.4 & \textbf{38.5} & 35.4 \\
\textbf{Filtered NQ} & 7.8 & 12.8 & 7.3 & 11.6 & \textbf{13.4} & 12.0 \\
\textbf{HotpotQA} & 11.2 & 16.1 & 13.4 & 10.7 & 14.4 & \textbf{16.1} \\
\bottomrule
\bottomrule
\end{tabular} }%
\end{spacing}
\caption{Results of LoRAMoE. Contrary to direct full fine-tuning and the use of LoRA-tuning that exhibits reduced performance on world knowledge benchmarks after training, our approach ensures simultaneous growth of both world knowledge benchmarks and other downstream tasks.}
\label{tab-moeresult}
\vspace{-0.5em}
\end{table*}

\section{Experiments}
\label{exps}
\subsection{Experiment Setup}

In this section, we introduce the training implementation for LoRAMoE. 
We only replace the linear layer in the feed-forward neural network of LLM with the LoRAMoE layer, initializing each layer with six experts, of which three experts are dedicated to addressing downstream tasks, and the other three are responsible for leveraging world knowledge in the base model by learning on its related tasks. The hyperparameters for control constraint strength $\beta$ and degree of imbalance $\delta$ are both set to $0.1$.
For LoRA settings, the $\alpha$, and $r$ are set to $32$ and four for the main result, respectively.
The dropout is $0.05$, and the learning rate is $2e-4$.
The training dataset is the 3 million set the same as the one described in Appendix~\ref{appendix:SFT-settings}, so as the evaluation settings.
We freeze the parameters of the base model, rendering only the experts and router in LoRAMoE trainable.
The batch size per node is set to $16$.

\subsection{Main Results}

Table~\ref{tab-moeresult} displays the performance of LoRAMoE and compares this result with the outcomes of directly applying SFT to the model or utilizing LoRA tuning. The results show that the language model with LoRAMoE gets good performance on both world knowledge benchmarks and others, indicating its effectiveness in avoiding knowledge forgetting while improving multi-tasking abilities.

For world knowledge benchmarks, contrary to the catastrophic collapse seen in Section \ref{sec:conflict}, LoRAMoE not only avoids this issue but also surpasses the model fine-tuned solely with the CBQA dataset. LoRAMoE shows a significant performance boost on world knowledge benchmarks over vanilla SFT, with up to a 63.9\% improvement and an average increase of 35.3\%. 



For other downstream tasks, LoRAMoE is capable of achieving performance close to or even surpassing that of direct SFT. For instance, in all reading comprehension tasks (i.e., Race, ReCoRD, multiRC), LoRAMoE achieved superior performance.

We also compare our method against PEFT by single LoRA. The knowledge forgetting also occurred during the single LoRA-tuning, as it is essentially the same as vanilla SFT \citep{hu2021lora}. Compared with a single LoRA, multiple collaborative LoRAs in LoRAMoE enhance both world knowledge retention and multitasking performance. They offer an average boost of 30.9\% in world knowledge benchmarks and 8.4\% in other downstream tasks.

Besides, $\mathcal{L}_{lbc}$ improves outcomes for LoRAMoE in the vast majority of tasks, both world knowledge benchmarks and others. Notably, for reading comprehension, NLI, and the original CBQA dataset, the benefits of this method were quite substantial, up to 17.6\%. This indicates capability partitioning in the expert group benefits the performance in multi-task learning.

\begin{table}[htbp]
  \centering
  \begin{spacing}{0.85}
    \setlength{\tabcolsep}{2.2mm}{
    \begin{tabular}{cccc}
    \toprule
    \toprule
    \textbf{\# Experts} & \textbf{\begin{tabular}[c]{@{}c@{}}\# LoRA\\ Rank\end{tabular}} & \textbf{\begin{tabular}[c]{@{}c@{}}\# Trainable\\ Param.\end{tabular}} & \textbf{\begin{tabular}[c]{@{}c@{}}Avg.\\ Results\end{tabular}} \\
    \midrule
    6         & 4         & 0.57\%             & 58.21        \\
    4         & 4         & 0.38\%             & 55.84        \\
    8         & 4         & 0.76\%             & 56.58        \\
    \midrule
    6         & 8         & 1.07\%             & 58.11        \\
    6         & 16        & 2.08\%             & 58.86       \\
    \bottomrule
    \bottomrule
\end{tabular} }%
\end{spacing}
  \caption{Performance of LoRAMoE varies with the number of experts and LoRA rank across all test sets. This includes the average results on both the world knowledge benchmark and all other downstream tasks. LoRAMoE shows stability to parameter changes.
  }
  \label{tab:sense-exp}
\end{table}

\subsection{Sensitivity Analysis}
\label{sec:sense-exp}

In this section, we analyze the parameter sensitivity of LoRAMoE. Keeping other settings constant, we vary the number of experts and the rank of LoRA. The average performance with varied parameter settings on all test sets including the world knowledge benchmark and all other downstream tasks is shown in Table~\ref{tab:sense-exp}. In Appendix~\ref{app:sense} there are detailed results.

As the number of trainable parameters increases, performance is generally stable. the number of 6 experts is the most beneficial choice, as more experts do not lead to higher performance. While the increase in LoRA rank improves the model's capabilities somewhat, it brings about an exponential rise in trainable parameters.

\subsection{Visualizing the Experts Utilization}

\begin{figure}
    \centering
    \includegraphics[width=0.48\textwidth]{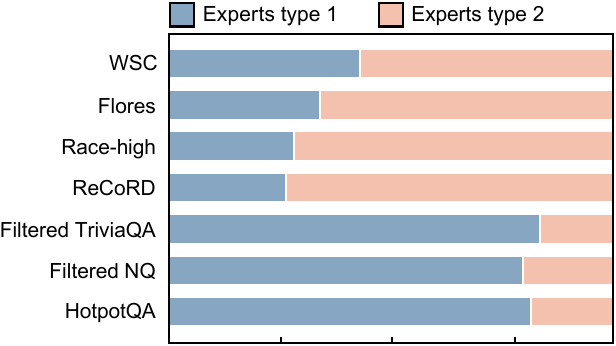}
    \caption{Visualization of routers' weight on different types of data, where type 1 refers to the experts dedicated to aligning the world knowledge in the base model with the human instruction and type 2 refers to the experts that focus on downstream tasks. The utilization rate of the type of experts diverged significantly across tasks.}
    \label{fig:router-vis}
\end{figure}

To confirm the effectiveness of LoRAMoE in specializing the experts with two types, we visualize their weight assigned by the router when encountered with data from downstream tasks and knowledge benchmarks respectively, as illustrated in Figure~\ref{fig:router-vis}.

There is a distinct contrast in the utilization of the two types of experts when dealing with world knowledge benchmarks and other downstream tasks. This suggests that the routers can automatically allocate specific tasks to experts with corresponding abilities during the inference phase.
Specifically, the experts requested to leverage world knowledge are greatly employed in world knowledge benchmarks (e.g., TriviaQA, Natural Questions, and HotpotQA), underscoring their vital role in preventing world knowledge forgetting. This corresponds to the fact we state in Section~\ref{sec:conflict} that supervised fine-tuning boosts the model’s capabilities in these tasks by associating pre-stored world knowledge in the model with human instructions. 
On the other hand, experts assigned to focus on enhancing performance in downstream tasks are given increased prominence when encountering these tasks. Through this visualized result, we find that some downstream tasks still require experts of another type. It is reasonable. For example, in reading comprehension tasks, the knowledge learned by the model during pre-training can better assist in making factual judgments. This phenomenon is even more pronounced in language-based tasks. In the WSC task \citep{levesque2012winograd}, the router allocates an average of about 45\% of its attention to the experts responsible for world knowledge.

\section{Related Work}

\textbf{Parameter-Efficient Fine-tuning.} 
With the size of language models growing larger, parameter-efficient fine-tuning (PEFT \citep{peft}) has become crucial for resource savings.
Researchers have proposed several approaches such as LoRA \citep{hu2021lora}, adapters \citep{adapters}, and prompt learning \citep{prompt_learnings}, to enhance fine-tuning efficiency. 
PEFT based on low-rank adapters \citep{hu2021lora} is popular and widely used, which introduces two trainable low-rank matrices in each fully connected layer, to achieve significant savings in training resources without adding additional inference computation cost. 
We apply low-rank techniques to the structure of experts to save resource consumption.


\textbf{Mixture-of-Experts.} 
The mixture of Experts (MoE) replaces the feed-forward neural network layer with sparsely activated experts, which significantly enlarges the model without remarkably increasing the computational cost \citep{moe1991first}.
Currently, the token-level MoE architectures are widely used in pre-trained language models and vision models \citep{shazeer2016outrageously, gshard, glam, riquelme2021scaling}.
In addition, researchers \citep{moe2022routing, chi2022representation} aim to investigate the router selection problem in MoE. 
Unlike these efforts to expand the model size and address the selection problem, we propose an MoE-style framework for multi-task learning and maintaining the world knowledge stored in LLMs.



\textbf{Multi-LoRA Architecture.}
Researchers also have utilized multiple LoRAs for enhanced model performance. 
\citet{huang2023lorahub} propose LoraHub to choose different LoRA combinations for task generalization. 
MOELoRA \citep{liu2023moelora} leverage LoRA and MoE for task-specific tuning and multitasking, especially in healthcare.
However, these methods need the data type as the input during the inference phase, which limits the application of the model to other tasks.
\citet{chen2023punica} first introduces multiple LoRA serving systems and \citet{sheng2023s} proposes S-LoRA, a system that can serve thousands of LoRA adapters from a single machine.
\citet{chen2023octavius} introduces several experts to enhance the model's ability for multimodal learning.
Unlike these approaches, LoRAMoE introduces an MoE-style plugin and Localize Balancing Constraint to tackle world knowledge forgetting in LLMs, while enhancing the model's ability to multi-task learning.

\section{Conclusion}
In this paper, we first delve into the conflict between improving LLM's performance on downstream tasks by scaling up data during the SFT phase and discouraging world knowledge forgetting.
To address this conflict, we then introduce LoRAMoE, a novelty framework for SFT, which introduces LoRAs as experts and integrates them by the router.
Extensive experimental results demonstrate that LoRAMoE can foster collaboration among experts to enhance the model's performance of downstream tasks, while preserving the world knowledge inside it.

\section{Limitations}
In this section, we discuss the potential limitations of our proposed method LoRAMoE. Firstly, although we have demonstrated the effectiveness of LoRAMoE in alleviating world knowledge forgetting while enhancing the downstream ability of the LLMs with SFT, we limit the model size to 7B due to resource and time constraints. Further work will be conducted on the larger LLMs, to understand the influence of large-scale SFT on these LLMs and to boost their multitasking abilities. Secondly, the localized balancing constraint can softly constrain the type of experts and balance the experts utilization. However, we haven't studied the case where there are more experts types for a more fine-grained task category. Future work will be conducted on a more fine-grained understanding of the influence of SFT and the utilization of LoRAMoE. 



\bibliography{custom}

\appendix
\section{Details about Experiment Implementation}
\label{appendix:SFT-settings}

\begin{table*}[htbp]
  \centering
  \begin{spacing}{0.9}
    \setlength{\tabcolsep}{2.5mm}{
\begin{tabular}{l|ccc}
\toprule
\toprule 
\textbf{Task Name}             & \textbf{\# Train} & \textbf{\# Test} & \textbf{Task Type}                                                                                                                                                                                              \\
\midrule
\textbf{TriviaQA} \citep{han2019episodic}   & 78785             & 254              & closed-book QA \\
\textbf{NQ} \citep{kwiatkowski2019natural} & 104071            & 357              & closed-book QA                                                                                          \\
\textbf{HotpotQA} \citep{qi2019answering}        & 72798             & 5622             & closed-book QA                                                                                                                               \\
\textbf{WSC} \citep{levesque2012winograd}              & 554              & 146             & coreference resolution         
\\
\textbf{WinoGrande}   \citep{sakaguchi2021winogrande}     & 40398             & 1767             & coreference resolution                                                                                            \\
\textbf{Flores}      \citep{guzman2019flores}      & 0                 & 1600             & machine translation                                                                                                                            \\
\textbf{WMT} \footnote{https://huggingface.co/datasets/wmt14, https://huggingface.co/datasets/wmt16}             & 500000            & -                & machine translation                                                                                                                                                                           \\
\textbf{RTE}      \footnote{https://aclweb.org/aclwiki/Recognizing\_Textual\_Entailment}         & 2490              & 3000             & NLI                                                                                             \\
\textbf{ReCoRD}    \citep{zhang2018record}        & 100730            & 10000            & reading comprehension                                                                                                                \\
\textbf{AX-g}      \footnote{https://super.gluebenchmark.com}        & 0                 & 356              & NLI                                                                          \\
\textbf{multiRC}  \citep{khashabi2018looking}         & 27243             & 9693             & reading comprehension                                                                             \\
\textbf{anli r1/r2/r3} \citep{liu2020microsoft}    & 162874            & -                & NLI                                                                \\
\textbf{qqp}   \citep{wang2017bilateral}            & 363846            & -                & NLI                                                                                                                                \\
\textbf{Xsum} \citep{narayan2018don}             & 204045            & 11334            & single-document summarization                                                                                                                            \\
\textbf{Race}   \citep{lai2017race}           & 87866             & 4934             & reading comprehension     \\
\textbf{duorc-selfRC}  \citep{saha2018duorc}   & 60721             & -                & reading comprehension                                                                                   \\
\textbf{AG-news}  \citep{zhang2015character}        & 120000             & -                & topic classification                                                                                                                                                                                      \\
\textbf{yelp review}  \citep{zhang2015character}    & 650000            & -                & sentiment classification                                                                                                                                                                                  \\
\textbf{openai tldr} \footnote{https://github.com/openai/summarize-from-feedback}     & 232188            & -                & summarization     \\
\bottomrule
\bottomrule
\end{tabular} }%
\end{spacing}
\caption{Details about the tasks in our fine-tuning dataset. "-" means we do not use the test set of this dataset for evaluation.}
\label{tab-datasets}
\end{table*}

\textbf{Datasets. } The seven tasks are closed-book question answering (CBQA), coreference resolution, natural language inference (NLI), abstract summarization, multi-lingual translation, reading comprehension, and text classification. 
Table \ref{tab-datasets} shows the composition of the 3-million-sample dataset.  The five million fine-tuning data we use includes three million versions and their variants from data augmentation strategies. The 1-million-sample version is the subset of the original 3-million-sample dataset.


\textbf{Evaluation. } We utilize the opencompass\footnote{\url{https://opencompass.org.cn/}} framework to run the evaluation process on the aforementioned tasks. Notably, considering previous work that has noted train-test overlap in CBQA datasets \citep{lewis2020question}, we elaborately select parts of the CBQA dataset without train-test overlap for our testing set, namely \textit{Filtered NQ} and \textit{Filtered TriviaQA}, to analyze the world knowledge of models better.


\section{The World Knowledge of LLM Further Declines after Being Trained with More Data}
\label{sec-1kw}
With the task types increasing, there is an inevitable trend to increase the amount of SFT training data. To further verify that a large-scale SFT training process can lead to knowledge forgetting of LLM as stated in Section~\ref{sec:conflict}, we construct a much larger dataset containing ten million training samples. In addition to the dataset from the previous section, we also added the following tasks:
\begin{itemize}
    \item Named Entity Recognition: sampled from \citet{wang2023instructuie}. Contains 17 different NER tasks.
    \item Program Execution: sampled from \citet{wang2022super}. Contains 90 different tasks requiring the LLM to understand the instructions about a program and execute it.
    \item Question Generation: sampled from a existing huggingface dataset \footnote{\url{https://huggingface.co/datasets/qa_zre}}. Given a context, the LLM needs to generate an appropriate question based on the answer.
    \item Text2sql: sampled from two existing huggingface datasets\footnote{\url{https://huggingface.co/datasets/Clinton/Text-to-sql-v1}, \url{https://huggingface.co/datasets/cfq}}. Given a description in natural language, the LLM needs to generate an appropriate sequence of SQL.
    \item Toxic Classification: sampled from a existing huggingface datasets\footnote{\url{https://huggingface.co/datasets/google/civil_comments}}.
\end{itemize}

After training the LLaMa-2-7b on this 10-million-sample dataset with the same experiment setup with Appendix~\ref{appendix:SFT-settings}, we find the LLM exhibit a greater knowledge-forgetting but a promising performance in other tasks apart from knowledge benchmarks.

\begin{table}[]
  \centering
  \begin{spacing}{0.8}
    \setlength{\tabcolsep}{2.5mm}{
\begin{tabular}{c|cc}
\toprule
\toprule
\textbf{Task Name}      & \textbf{Baseline} & \textbf{Result} \\
\midrule
\textbf{NER}                     & 42.1              & 82.2            \\
\textbf{Program Execution}        & 18.7              & 78.5            \\
\textbf{Toxic Classification}              & 96                & 97.4            \\
\textbf{Question Generation}      & 46.2              & 61.1            \\
\textbf{Text2sql}            & 56                & 96.2            \\
\textbf{WSC}             & 65.4              & 70.2            \\
\textbf{winogrande}             & 61.7              & 66.1            \\
\textbf{Flores}                  & 0.1               & 26.0            \\
\textbf{Xsum}                   & 19.7              & 33.2            \\
\textbf{Race-middle}            & 30.5              & 87.0            \\
\textbf{Race-high}                & 30.4              & 83.3            \\
\textbf{RTE}                      & 52.7              & 87.4            \\
\textbf{ReCoRD}                   & 29.5              & 56.6            \\
\textbf{AX-g}                     & 52.0              & 87.9            \\
\textbf{multiRC}                  & 44.0              & 86.0            \\
\midrule
\textbf{TriviaQA}                & 52.2              & 30.9            \\
\textbf{NQ}         & 18.5              & 14.2            \\
\textbf{Filtered TriviaQA} & 33.5              & 15.7            \\
\textbf{Filtered NQ}          & 7.8               & 5.0   
\\
\textbf{HotpotQA}                & 11.2              & 7.6              \\

\bottomrule
\bottomrule
\end{tabular} }%
\end{spacing}
\caption{Performance of Llama-2-7B after vanilla SFT with a 10-million-sample datasets. There is a much more severe decrease in the performance on the CBQA tasks, while a great enhancement in other tasks compared with the baseline. }
\end{table}

\section{Mixed Distribution Dilemmas for Expert Balancing}
\label{app-balance}

When fine-tuning MoE without any constraints, the router mechanism often converges to a state in which a small number of experts receive a disproportionately large share of preferences by the router, as depicted in Figure~\ref{fig:imbalance_loss}.
This imbalance among experts presents a challenge to correct, as experts that receive greater routing weights in the early stages of training undergo more rapid optimization, thereby garnering increased preferences from the router. 
A similar phenomenon has been documented in the work presented in \citet{shazeer2016outrageously} and \citet{switchtransformers}.

A conventional solution for balancing experts utilization involves employing the coefficient of variation of the experts' importance as the loss function, aimed at equalizing the significance of each expert \citep{shazeer2016outrageously}.
This solution assumes that the distribution of training samples for optimising MoE is a single distribution, which inherently eliminates the necessity of considering the diverse origins of data distribution.
Specifically, this traditional approach simplifies the modeling process by assuming homogeneity in data sources that often do not align with fine-tuning data containing both factual knowledge QA and other downstream tasks. 
Therefore, such simplification can lead to significant biases, particularly when encountering datasets with varied distributional characteristics.

Traditional balancing constraints, which aim to allocate a uniform distribution of training samples across all experts, can lead to inaccurate parameter estimation. This is because such constraints do not account for the intrinsic differences in data representation and importance across various categories.  Recognizing the disparate nature of data distributions, LoRAMoE strategically assigns data to experts, not uniformly, but based on the observed imbalances. This allocation is governed by a set of weights that are calibrated to reflect the varying significance and representation of different data categories within the overall dataset.

Such a specialized allocation method is pivotal in addressing the challenges posed by uneven data distributions. By tailoring the distribution of training samples to each expert based on the inherent disparities in the data, LoRAMoE facilitates a more accurate and representative parameter estimation. This nuanced approach to data distribution allows for a more effective fitting of the model to diverse data subsets, significantly enhancing the model's predictive accuracy and generalization capability. This strategy is particularly effective in scenarios where data imbalance could otherwise lead to skewed learning and generalization errors, ensuring that each data category is appropriately represented and modeled within the overall system.

To illustrate the concept with a simplified model, let's assume our training data is sampled from a mixture of two Gaussian distributions. The means $(\mu_1, \mu_2)$ and variances $(\sigma_1^2, \sigma_2^2)$ of these distributions are implicit. The proportion of training data from each distribution is denoted as $p_1 \text{and} P_2$ where $p_1+p_2=1$, without loss of generality, we assume that $p_1 \leq p_2$. When a MoE model fits the proposed distribution with balanced weights $m$, the likelihood of the model given the data can be expressed as:

{\small
\begin{align}
L(\mathbf{X}) = & \prod_{x \in \mathbf{X}_1}\left(m \mathcal{N}\left(x ; \mu'_{1}, \sigma'^{2}_{1}\right)\right. \nonumber \\
                & \left. + (1-m) \mathcal{N}\left(x ; \mu'_{2}, \sigma'^{2}_{2}\right)\right) \nonumber \\
                & \times \prod_{x \in \mathbf{X}_2}\left(m \mathcal{N}\left(x ; \mu'_{1}, \sigma'^{2}_{1}\right)\right. \nonumber \\
                & \left. + (1-m) \mathcal{N}\left(x ; \mu'_{2}, \sigma'^{2}_{2}\right)\right),
\end{align}
}

\begin{table*}[htbp]
  \centering
  \begin{spacing}{0.9}
    \setlength{\tabcolsep}{2.5mm}{
\begin{tabular}{ccccc}
\toprule
\toprule
\textbf{Task Name}          & \textbf{\begin{tabular}[c]{@{}c@{}}\# Expert=8\\      \# rank=4\end{tabular}} & \textbf{\begin{tabular}[c]{@{}c@{}}\# Expert=4\\      \# rank=4\end{tabular}} & \textbf{\begin{tabular}[c]{@{}c@{}}\# Expert=6\\      \# rank=8\end{tabular}} & \textbf{\begin{tabular}[c]{@{}c@{}}\# Expert=6\\      \# rank=16\end{tabular}} \\
\midrule
\textbf{WSC}                & 71.2                                                                          & 76.0                                                                          & 70.2                                                                          & 76.9                                                                           \\
\textbf{winogrande}         & 69.8                                                                          & 56.0                                                                          & 69.5                                                                          & 70.9                                                                           \\
\textbf{Flores}             & 25.0                                                                          & 25.8                                                                          & 26.1                                                                          & 26.3                                                                           \\
\textbf{Xsum}               & 32.8                                                                          & 33.3                                                                          & 33.7                                                                          & 34.0                                                                           \\
\textbf{Race-middle}        & 90.3                                                                          & 84.2                                                                          & 90.3                                                                          & 90.5                                                                           \\
\textbf{Race-high}          & 87.1                                                                          & 80.7                                                                          & 87.3                                                                          & 87.2                                                                           \\
\textbf{RTE}                & 84.5                                                                          & 80.1                                                                          & 88.1                                                                          & 85.2                                                                           \\
\textbf{ReCoRD}             & 85.6                                                                          & 85.5                                                                          & 86.0                                                                          & 86.1                                                                           \\
\textbf{AX-g}               & 88.8                                                                          & 77.5                                                                          & 88.2                                                                          & 85.7                                                                           \\
\textbf{multiRC}            & 77.2                                                                          & 87.6                                                                          & 81.1                                                                          & 87.3                                                                           \\
\midrule
\textbf{TriviaQA}        & 54.4                                                                          & 57.8                                                                          & 58.2                                                                          & 58.9                                                                           \\
\textbf{NQ}                 & 25.6                                                                          & 27.9                                                                          & 27.8                                                                          & 28.2                                                                           \\
\textbf{Filtered TriviaQA} & 30.7                                                                          & 35.8                                                                          & 36.7                                                                          & 34.3                                                                           \\
\textbf{Filtered NQ}        & 11.5                                                                          & 13.4                                                                          & 12.0                                                                          & 15.4 \\
\textbf{HotpotQA}          & 14.5                                                                          & 16.0                                                                          & 16.4                                                                          & 16.5                                                                           \\

\bottomrule
\bottomrule
                                                                
\end{tabular} }%
\end{spacing}
      \caption{Detailed result on sensitivity study.
  }
  \label{tab:app-sense-exp-det}
\end{table*}

where $Card( \mathbf{X}_1):Card(\mathbf{X}_2) = p_1:p_2$. Using $\mathcal{N}_1(x)$ and $\mathcal{N}_2(x)$ for $ \mathcal{N}\left(x ; \mu^{'}_{ 1}, \sigma^{'2}_{ 1}\right)$ and $ \mathcal{N}\left(x ; \mu^{'}_{ 2}, \sigma^{'2}_{ 2}\right)$,

The optimal mean value for $\mu^{'}_{ 1}$ satisfies the following conditions, whose value is 0 when the fitted distribution is in the same family of mixed distributions $\mathbb{N}(\theta, p_1)$ as the sampling distribution:
{\small
\begin{align}
    \frac{\partial \log L(\mathbf{X})}{\partial \mu_1^{\prime}} = & \sum_{x \in \mathbf{X}_1 \cup \mathbf{X}_2} \frac{\partial}{\partial \mu_1^{\prime}} \log \left(m \mathcal{N}_1(x) \right. \nonumber \\
    & \left. + (1-m) \mathcal{N}_2(x)\right) \nonumber \\
    = & \sum_{x \in \mathbf{X}_1 \cup \mathbf{X}_2} \left(\frac{x-\mu_1^{\prime}}{\sigma_1^{\prime 2}}\right) \nonumber \\
    & \times \frac{m \mathcal{N}_1(x)}{m \mathcal{N}_1(x) + (1-m) \mathcal{N}_2(x)},
\end{align}
}

In equation 10, we can replace part of the summation with the empirical estimate of the mean of the input x. For an ideal routing network, there must exist a distribution $N_i$ such that the data allocated to this distribution is independently and identically distributed with one of the peaks in the sampling distribution. Let's assume this distribution to be $N_2$. In this case, if $m \geq p_1$, then the fitting result for distribution $\mu_{1'}$ will be $\mu_1' = (p_1 \mu_1 + (m-p_1) \mu_2) / m$. Based on the chain rule of differential derivation, we end up with:
{\small
\begin{align}
    \frac{d \log L}{d m} =& \frac{\partial \log L}{\partial \mu_1^{\prime}} \frac{d \mu_1^{\prime}}{d m} \nonumber \\
    =& \left( \sum_{x \in \mathbf{X}_1 \cup \mathbf{X}_2} \left(\frac{x-\mu_1^{\prime}}{\sigma_1^{\prime 2}}\right) \right. \nonumber \\
    &\quad \left. \times \frac{m \mathcal{N}_1(x)}{m \mathcal{N}_1(x) + (1-m) \mathcal{N}_2(x)} \right) \nonumber \\
    &\quad  \times \frac{p_1 (\mu_2- \mu_1)}{m^2}  \nonumber \\
    \leq& 0,
\end{align}
}

The inverse result can be derived similarly. Therefore, the best training error is achieved only when the mixing coefficient $m$ of the prior distribution is consistent with the actual sampling distribution weight $p_1$.

\section{Detalied Results of Sensitivity Study}
\label{app:sense}
Table~\ref{tab:app-sense-exp-det} shows the detailed results presented in Section~\ref{sec:sense-exp}.

\end{document}